%% file: main.tex
\documentclass[10pt,twocolumn,letterpaper]{article}
\pdfoutput=1

\usepackage[pagenumbers]{cvpr}

\usepackage[dvipsnames]{xcolor}
\definecolor{cvprblue}{rgb}{0.21,0.49,0.74}
\usepackage[pagebackref,breaklinks,colorlinks,citecolor=cvprblue]{hyperref}

\hypersetup{
    pdfauthor={Qilong Zhangli},
    pdftitle={Layout-Agnostic Scene Text Image Synthesis with Diffusion Models},
    pdfsubject={Computer Vision, Artificial Intelligence},
    pdfkeywords={Scene Text Synthesis, Diffusion Models, Text Image Synthesis, Visual Text Generation},
}

\usepackage{multirow}
\usepackage{float}
\usepackage{algorithm}
\usepackage{algpseudocode}
\usepackage{amssymb}

\newcommand*{\affmark}[1][*]{\textsuperscript{#1}}

\title{Layout-Agnostic Scene Text Image Synthesis with Diffusion Models}

\author{
Qilong Zhangli\affmark[1,2]\quad 
Jindong Jiang\affmark[1]\quad 
Di Liu\affmark[1]\quad 
Licheng Yu\affmark[2]\quad 
Xiaoliang Dai\affmark[2]\quad \\
Ankit Ramchandani\affmark[2]\quad 
Guan Pang\affmark[2]\quad 
Dimitris N. Metaxas\affmark[1]\quad 
Praveen Krishnan\affmark[2]\\
{
\affmark[1]Rutgers University\quad
\affmark[2]Meta AI
}
}

\begin{document}

\maketitle
\input{sections/0_abstract}    
\input{sections/1_introduction}

\input{sections/2_related_work}

\input{sections/3_methodology}
\input{sections/4_experiments}
\input{sections/5_discussion_and_conclusion}
{
    \small
    \bibliographystyle{ieeenat_fullname}
    \bibliography{main}
}
\end{document}

%% file: sections/0_abstract.tex
\begin{abstract}
While diffusion models have significantly advanced the quality of image generation, their capability to accurately and coherently render text within these images remains a substantial challenge. Conventional diffusion-based methods for scene text generation are typically limited by their reliance on an intermediate layout output. This dependency often results in a constrained diversity of text styles and fonts, an inherent limitation stemming from the deterministic nature of the layout generation phase. To address these challenges, this paper introduces SceneTextGen, a novel diffusion-based model specifically designed to circumvent the need for a predefined layout stage. By doing so, SceneTextGen facilitates a more natural and varied representation of text. The novelty of SceneTextGen lies in its integration of three key components: a character-level encoder for capturing detailed typographic properties, coupled with a character-level instance segmentation model and a word-level spotting model to address the issues of unwanted text generation and minor character inaccuracies. We validate the performance of our method by demonstrating improved character recognition rates on generated images across different public visual text datasets in comparison to both standard diffusion based methods and text specific methods. 
\footnote{This arXiv resubmission of SceneTextGen, accepted to CVPR 2024, is for Google Scholar indexing. \href{https://openaccess.thecvf.com/content/CVPR2024/papers/Zhangli_Layout-Agnostic_Scene_Text_Image_Synthesis_with_Diffusion_Models_CVPR_2024_paper.pdf}{CVPR paper}}
\end{abstract}

%% file: sections/1_introduction.tex
\vspace{-6pt}
\section{Introduction}
\begin{figure}[t]\centering
    \includegraphics[width=1.0\linewidth]{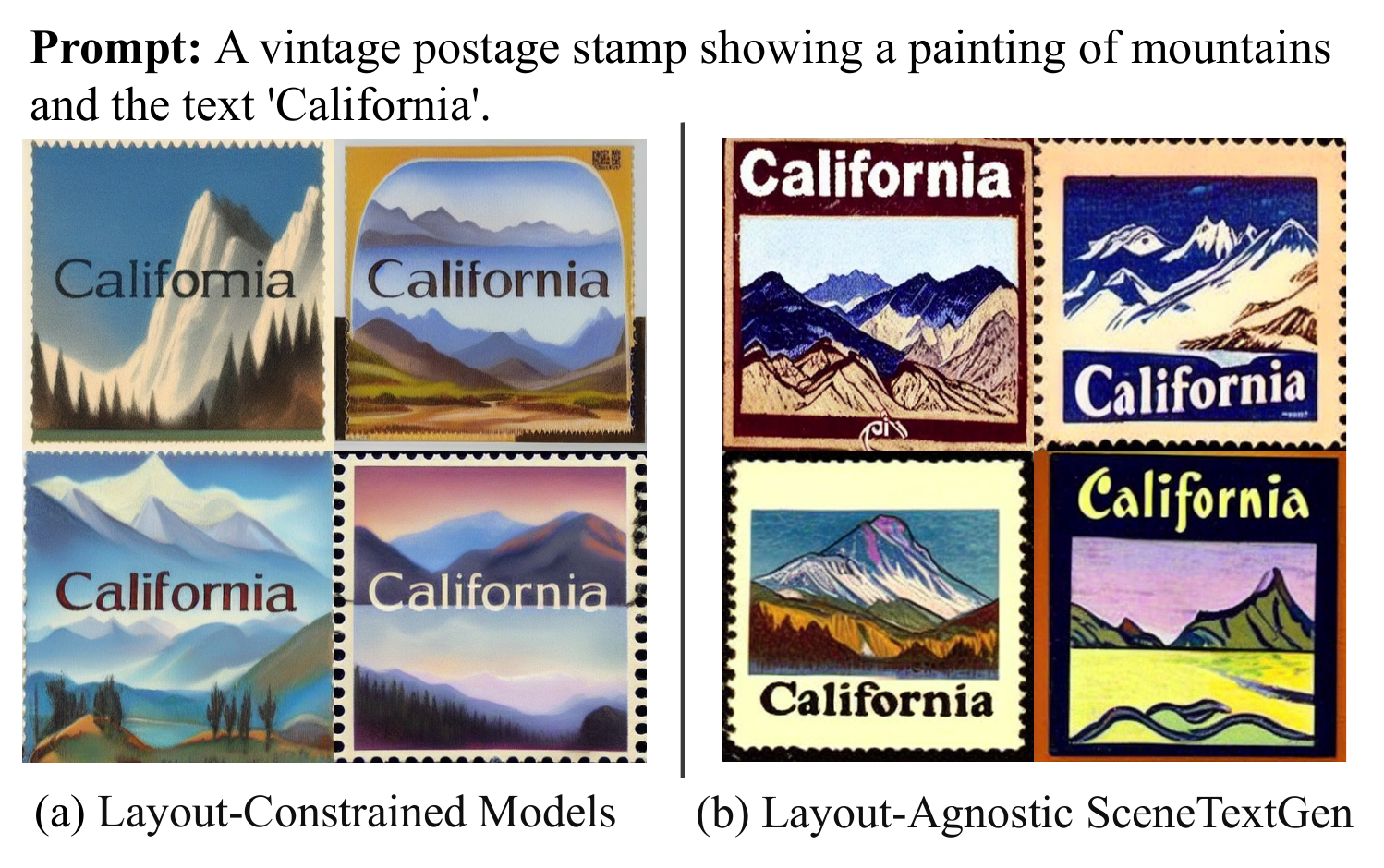}
    \caption{
Models reliant on predefined text layouts for input exhibit limitations such as constrained font diversity and static text positioning during each inference, leading to a lack of variability in style and arrangement.
        }
    \vspace{-10pt}
    \label{fig:difference}
\end{figure}

The text-to-image~\cite{tao2020deep, xu2018attngan} task has gained popularity with advancements in diffusion models~\cite{Imagen, DALLE, deepFloyd-if}, significantly enhancing the quality of image generation. However, seamlessly integrating clear and contextually appropriate text into images is a persistent challenge. Text plays a vital role in many domains, including media content creation and artistic design, yet current diffusion models often struggle to produce text that is lexically correct, in valid font style, and that naturally complements the overall image aesthetics.

Traditional methods~\cite{zhan2019spatial,wu2019editing,yang2020swaptext} of creating scene text images typically formulate this problem as scene text editing which only involves editing or adding text to an existing scene image. These methods have challenges handling complex backgrounds, font styles and lighting variations. While recent models \cite{Imagen, ediffi, deepFloyd-if, TextDiffuser, characterAware} have made strides in addressing these limitations by enhancing text encoding strategies \cite{characterAware} or employing predefined text layouts \cite{TextDiffuser}, they still face significant constraints in generating visual text. These constraints become apparent when observing the limited diversity in font styles and the static positioning of text, as shown in Fig.~\ref{fig:difference}. Such rigidity in layout and font selection hampers the capacity of generative models to produce text that is stylistically varied and contextually aligned with the image content.

To address these issues, we propose \textbf{SceneTextGen}, a novel framework that capitalizes on the capabilities of latent diffusion models to infuse text into scene images with greater diversity and authenticity. 
Our approach is specifically engineered to transcend the limitations of predefined layouts, enabling more flexible text placement and an expansive assortment of text styles.

The two primary contributions of this work are: (i) the integration of character-level encoder to capture the typographic properties of visual text and carefully injecting it into the cross attention layers of the diffusion model, and (ii) introducing a novel word spotting loss using a pre-trained OCR along with a character segmentation loss to make the network more faithful in generating visual text. The character level encoder naturally blends with the existing diffusion model architecture and therefore allows the network to learn the layout of visual text implicitly rather than constraining itself to some pre-defined form. Our comprehensive evaluations confirm that SceneTextGen surpasses contemporary methods, facilitating the generation of images with text that is both aesthetically pleasing and rich in variety.

\input{figures/teaser}

%% file: figures/teaser.tex

\begin{figure*}[!htbp]\centering
    \vspace*{-2em}
    \includegraphics[width=1.0\linewidth]{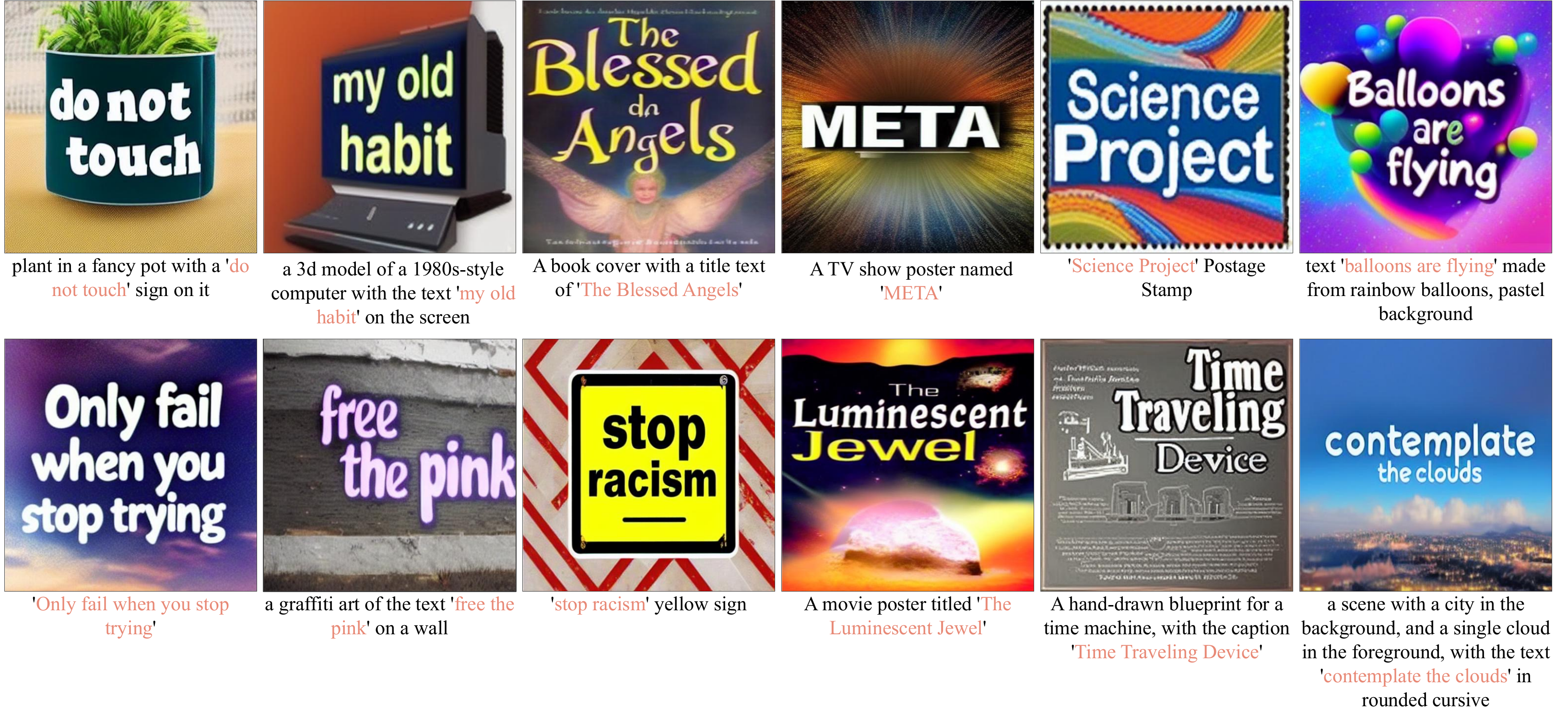}
    \vspace*{-2em}
    \captionof{figure}{\textbf{Scene Text Generation.} Qualitative samples of scene text images generated by our model are presented. These images contain visually appealing texts that are coherent with the background and are created without relying on any spatial information or predefined layouts as input, thereby enhancing the Text-to-Image (T2I) Diffusion Model's capability to generate text.}
    \label{fig:teaser}
\end{figure*}

%% file: sections/2_related_work.tex
\section{Related Work}

\subsection{Text 2 Image Generation Models}
In the realm of text-to-image generation, diffusion models~\cite{ho2022cascaded,controlNet,dreambooth,DALLE,deepFloyd-if,gu2022vector,zhang2023sine,li2023gligen,nichol2021glide} represent a significant leap forward. Different from GAN-based models~\cite{goodfellow2014generative,zhu2017unpaired,isola2017image}, diffusion models employ a stochastic process that iteratively adds noise to an image and learn to reverse this process to generate images from textual descriptions. Their capacity to produce high-quality, detailed visual content from text prompts has been well documented. 
The Latent Diffusion Model (LDM)\cite{stableDiffusion} further enhances this approach by operating in a compressed image latent space, improving both efficiency and image quality. It also facilitates high-quality conditional generation through cross-attention text conditioning, leading to various downstream applications \cite{prompt_to_prompt,Han_2024_WACV,han2023svdiff,jiang2023object,wu2023slotdiffusion}.
DALL-E \cite{DALLE}, renowned for its novel approach of combining discrete VAEs with transformer language models, has shown remarkable ability in generating diverse and complex images from textual descriptions, showcasing the potential of transformer architectures in creative generative tasks.
Deepfloyd \cite{deepFloyd-if}, utilizing the robust T5 text encoders \cite{T5}, not only enhances image quality but also facilitates nuanced understanding of complex prompts, thereby allowing for more accurate and context-aware visual representations.
ControlNet~\cite{controlNet}, has demonstrated exceptional performance in conditioned image generation by providing the model with references such as skeleton, canny edge images, and segmentation maps.
However, these models frequently encounter difficulties when it comes to the generation of text within images (see Fig. \ref{fig:comparison}), a task requiring the text to be not only visually integrated but also contextually pertinent to the image content. This challenge stems from the complexity of modeling the fine-grained interplay between visual and textual elements, ensuring that the generated text is legible, aesthetically fitting, and semantically in sync with the image. Prior attempts to refine this aspect have led to improvements~\cite{deepFloyd-if,TextDiffuser,controlNet}, yet the generation of contextually coherent text in images remains a largely unsolved problem, underscoring the need for more focused research.

\subsection{Scene Text Generation}
\label{subsec:stg}

The success of adversarial networks such as GANs~\cite{goodfellow2014generative,zhu2017unpaired,isola2017image,chang2022deeprecon,xia2022sign} for image generation and style transfer~\cite{karras2017progressive,karras2020analyzing} gave rise to scene text generation methods which can generate text at the granularity of glyphs~\cite{azadi2018multi,li2020fet} or individual words~\cite{wu2019editing,yang2020swaptext,krishnan2023textstylebrush}. Many previous works focused on the particular task of scene text editing where the model learns a style from a reference image and renders the target content in that style. Methods such as SRNet~\cite{wu2019editing}, SWAPText~\cite{yang2020swaptext} decomposes the problem into: (1) learning the foreground text using style and content, (2)  background in-painting network to remove existing text, and (3) a blending network to merge foreground and background. These methods often fail in learning the correct style and removing existing text from complex backgrounds. TextStyleBrush~\cite{krishnan2023textstylebrush} proposes a self-supervised approach to disentangle content and style and generate word images in a one-shot manner. All these previous methods are limited to generating individual words, requires reference word style images and does not generalize to generate the entire scene text image.

Spatial fusion GAN (SF-GAN)~\cite{zhan2019spatial} generates text images by superimposing a foreground content image, transformed to match the style and geometry of a background image. This approach is aimed more at pure text synthesis on image regions without text, whereas in this work we aim to generate both image and text in a manner that reflects its natural appearance in real-world contexts.

With the significant progress of diffusion models in text-to-image generation, recent methods in scene text generation adapt these models towards producing more legible visual text. DIFFSTE~\cite{diffSTE} enhances scene text editing with a dual encoder design in diffusion models, promoting text legibility and style control. It is adept at mapping text instructions to images, showcasing zero-shot capabilities for rendering text in novel font variations and interpreting informal natural language instructions. This method is however a scene text editing method which generates single keywords on specific regions defined by a mask, whereas we propose a scene text generation network. One of the closest work in this space is 
TextDiffuser~\cite{TextDiffuser} which address the problem of scene text generation by decomposing the problem into two-stages. In the first stage, a transformer model create the layout mask for keywords extracted from text prompts. This is then taken as condition data while formulating the diffusion model to generate scene text images. They also introduce the character segmentation loss which helps in generating legible text.  
ControlNet~\cite{controlNet}, though not originally designed for scene text image rendering, has been effectively adapted for this purpose. It utilizes canny edge maps as conditional inputs, sourced from printed text images generated by a layout model, to fine-tune the diffusion model's output. The use of pre-defined layout for visual text generation in~\cite{TextDiffuser, controlNet}, seriously limits the diversity of text styles, fonts and even the layouts. We believe this is due to the inherent difficulty in predicting layouts independently without the general image guidance. In our work, we avoid the need of pre-defined layout and make the network implicitly learn layout along with image generation using our novel way of injecting character level features. We also introduce a word spotting loss which augments the character segmentation loss proposed in~\cite{TextDiffuser} to generate more legible visual texts.

\subsection{Scene Text Recognition}
Advances in computer vision have laid a foundation for sophisticated analytical techniques in various domains ~\cite{baek2019wrong,liu2024lepard,liu2023deformer,gao2022data,martin2023deep,he2023dealing,liu2022transfusion,zhangli2022region,sayadi2022harnessing,ye2023free,ye2023demultiplexing,wen2024second}. 
Especially in scene text image recognition, most existing works split the process into two stages: a text detection~\cite{ma2018arbitrary,litman2020scatter,baek2019character,liao2020real,zhou2017east,shi2018aster} module to detect words or characters from complex backgrounds, and a text recognition~\cite{baek2019wrong,wang2022multi,lee2016recursive, shi2016robust} module which transcribes the text into unicode characters given a cropped word image. More recently, end-2-end methods~\cite{buvsta2019e2e,qin2019towards,li2023trocr,wang2021towards,glassOCR} have become popular due to the benefits of joint training of detector and recognizer to share contextual information. 

In the nexus of scene text generation and recognition, leveraging pre-trained scene text recognition or word spotting models as guidance during diffusion-based text-to-image synthesis has surfaced as an innovative strategy. For simplicity, in this paper we refer to a scene text recognition module as OCR (optical character recognition). The integration of OCR-derived losses enables the refinement of generative models to produce text that is not just visually coherent but also contextually accurate. This confluence of generative modeling prowess with OCR accuracy paves the way for novel research avenues to generate images with text that is both authentic to read and visually integrated.

%% file: sections/3_methodology.tex
\section{Methodology}

\subsection{Motivation}

Our objective is to facilitate layout-free text image generation with diverse layouts and styles. A straightforward approach would be to directly fine-tune an existing latent diffusion model with text images. However, our early investigation suggests that this strategy did not yield significant improvements compared to the original LDM. We hypothesize that this limitation is due to two primary factors. First, the language encoder of the LDM, primarily designed for semantic interpretation, fails to capture character-specific information adequately for text rendering. This encoder tends to provide more semantic than structural information about the text, necessitating a dedicated network for encoding the text and guiding the model on its visual representation. Second, the conventional denoising loss used in diffusion models seems insufficient for accurately rendering text in images, often leading to text regions resembling textual patterns without the distinct features of text strokes. To address these challenges, we propose integrating a character-level encoder and a hierarchical cross-attention mechanism to learn character-level context information. Additionally, we introduce two auxiliary losses at the word and character levels to emphasize the text presentation. The subsequent sections will detail each of these components and their contribution to enhancing our model's performance.

\begin{figure*}[!htbp]\centering
    \includegraphics[width=1.0\linewidth]{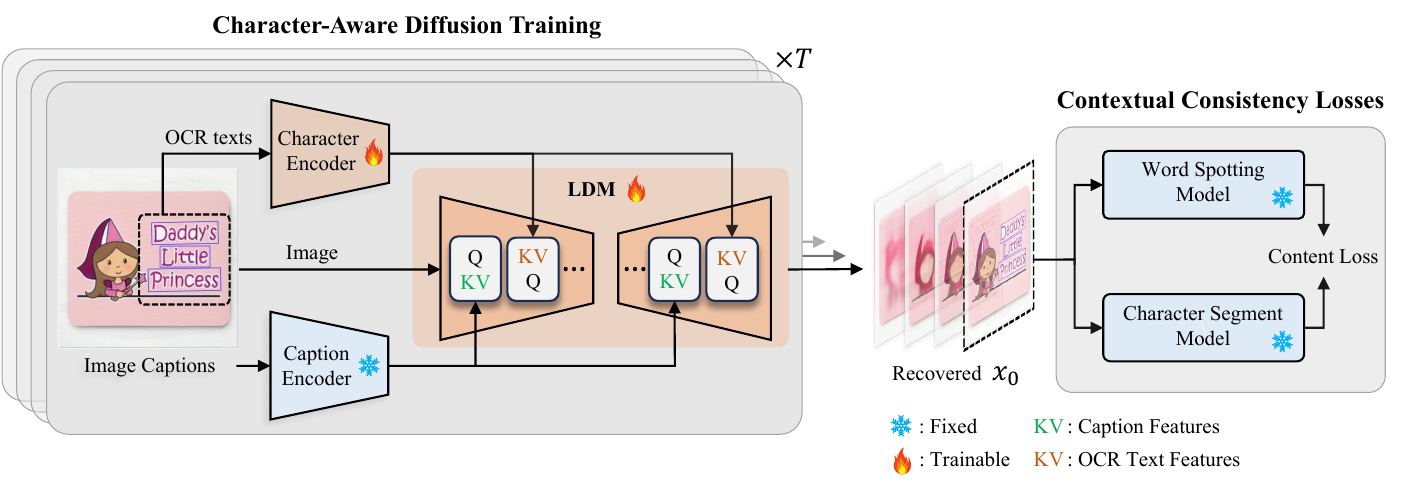}
    \caption{
Model Framework: SceneTextGen employs a character-level encoder to extract detailed character-specific features. During loss computation, the model leverages both word-level and character-level supervisions to guide the recovery of the image, in addition to the standard denoising loss. This dual-level supervision enhances the model's ability to accurately generate and refine text within scenes. 
        }
    \label{fig:method_overview}
\end{figure*}

\subsection{Preliminaries}

The proposed method is trained on a corpus of visual text images. Given an image $\mathcal{I}$ with textual caption $c$, we denote $w_{i}$, where $i \in [0, N]$, as the visual text (words) present in this image. We also assume that for each word, we know its location, given in terms of the bounding box. For all practical purposes, we assume this can be pre-computed using a pre-trained OCR detection~\cite{liao2020real} and recognition network~\cite{bautista2022scene}. Please note, the OCR bounding box and transcriptions are typically noisy. Fig.~\ref{fig:method_overview} presents an overview of our proposed method. A latent diffusion model has three key components: a CLIP encoder for semantic embedding of text and images, an autoencoder for dimensionality reduction and feature extraction, and a UNet-based structure for effective image-text synthesis and manipulation. The core architecture of our method follows the latent diffusion model, which is adapted to incorporate the proposed character-level features and content-based loss functions.

\subsection{Character-Level Encoding}
Our approach begins by extracting and ordering the visual text words from the image based on its locations as given by the bounding box. 

We sort the word boxes following the conventional reading pattern, i.e., from left to right and top to bottom. 
This provides us the naive ordering of visual text which is then tokenized at the character level to capture its typographic information. Then, a character encoder is used to encode these tokenized characters into a high-dimensional feature space, allowing an accurate transcription of the text's spelling and appearance. The derived text features encapsulate the typographical details and are poised for the subsequent integration with the image's latent features, which have already been contextually primed by the initial cross-attention with the encoded caption features obtained from the CLIP~\cite{radford2021learning} text encoder.

\subsection{Hierarchical Text Integration Process}
We have employed a sequential cross-attention mechanism to integrate character information into the U-Net architecture effectively. This method is inspired by the observation that the cross-attention of captioning features is pivotal in shaping the overall spatial structure of an image, a notion supported by prior studies \cite{prompt_to_prompt}. Leveraging this concept, our model initially constructs the general spatial layout of the image content, guided by the caption, using the first cross-attention layer. It then focuses on the precise rendering of characters in a subsequent cross-attention phase. This approach establishes a hierarchical text integration process, facilitating the development of a preliminary visual scaffold that is both thematically and contextually coherent. It ensures that the textual elements are accurately positioned and seamlessly integrated into the overall image structure.

\subsection{OCR-Guided Diffusion for Text Accuracy}
To ensure the textual accuracy of generated images, our model incorporates an OCR loss, utilizing the predictions from the end to end pre-trained GLASS model~\cite{glassOCR}. Upon each iteration of the diffusion process, the UNet predicts a denoising step, from which we derive an estimation of clean image \( x_0 \). This estimated \( x_0 \) is then decoded through a Variational Autoencoder (VAE)~\cite{vae} to reconstruct an image.

The GLASS OCR model performs inference on this reconstructed image to produce word-level recognition results, which are represented as a tensor \( P \) with shape \([N, L, K]\), where \( N \) is the number of words detected, \( L \) is the maximum length of any word, and \( K \) is the size of the character set. The ground truth for these detections is represented as a tensor \( G \) with shape \([N, L]\), where each entry is the character index for the corresponding position, and non-character positions are marked with zero.

The OCR loss (\( \mathcal{L}_{\text{OCR}} \)) is computed using a masked cross-entropy function, which is formulated as follows:

\begin{equation}
\mathcal{L}_{\text{OCR}} = -\frac{1}{\sum M} \sum_{i=1}^{N} \sum_{j=1}^{L} M_{ij} \cdot \log \left( \frac{\exp(P_{ij}[G_{ij}])}{\sum_{k=1}^{K} \exp(P_{ij}[k])} \right)
\end{equation}

Here, \( M \) is a binary mask tensor that has the same shape as \( G \) and indicates the valid character positions (i.e., where \( G_{ij} \neq 0 \)). \( M_{ij} \) represents the binary value of the mask at the \( i \)-th word and \( j \)-th character position, \( P_{ij} \) is the predicted probability distribution over the character set, and \( G_{ij} \) is the ground truth character index at that position.

By integrating this OCR loss into the training regime, we guide the diffusion model to produce text that is not only visually coherent but also textually accurate, as recognized by the GLASS~\cite{glassOCR} OCR model, thereby enhancing the overall fidelity of the generated images.

\subsection{Refinement of Text Generation with Character-Level Constraints}
During the iterative refinement of our diffusion model, we observed an unintended consequence of the OCR loss; the model tended to generate images with repetitive words. This issue is potentially attributed to the blurry nature of images at higher noise levels during training, which could render the OCR loss counterproductive. The word-level OCR loss, while ensuring textual accuracy, imposes no explicit constraint on the quantity of text within the image, inadvertently encouraging the model to generate excessive text.

To address this, we augmented our loss function with a character-level segmentation loss, which acts directly on the latent space rather than the recovered image. After obtaining the predicted latent features of a image \( x_0 \) from the UNet, we proceed in two directions: we decode \( x_0 \) using the VAE to compute the word-level OCR loss on the recovered image (as explained earlier), and we also apply \( x_0 \) to a pre-trained character-level segmentation model based on U-Net adapted from ~\cite{TextDiffuser}. This model outputs a 96-dimensional feature map (corresponding to the length of the alphabet plus one for non-character pixels) with a spatial resolution of 64×64. The character-aware loss is then computed via cross-entropy between this feature map and a resized character-level segmentation mask \( C \).

Thus, the total loss function is a composite of the denoising loss, the word-level recognition loss, and the character-level segmentation loss, expressed as:

\begin{equation}
\mathcal{L}_{\text{total}} = \mathcal{L}_{\text{denoising}} + \lambda_{\text{word}} \mathcal{L}_{\text{OCR-word}} + \lambda_{\text{char}} \mathcal{L}_{\text{OCR-char}}
\end{equation}

where \( \mathcal{L}_{\text{denoising}} \) is the denoising loss, \( \mathcal{L}_{\text{OCR-word}} \) is the word-level OCR loss, \( \mathcal{L}_{\text{OCR-char}} \) is the character-level segmentation loss, and \( \lambda_{\text{word}} \) and \( \lambda_{\text{char}} \) are weighting coefficients balancing the contribution of each term.

The character-level segmentation loss is formulated as:

\begin{equation}
\mathcal{L}_{\text{OCR-char}} = -\frac{1}{HW} \sum_{h=1}^{H} \sum_{w=1}^{W} \sum_{c=1}^{96} y_{hwc} \log \left( \hat{y}_{hwc} \right)
\end{equation}

where \( H \) and \( W \) are the height and width of the feature map, \( y \) is the ground truth character segmentation mask, and \( \hat{y} \) is the predicted character probability map from the segmentation model.

By integrating character-level information directly in the latent space, we impose a structured constraint on text generation, promoting both the accuracy and the appropriate quantity of text in the generated images.

%% file: sections/4_experiments.tex
\section{Experiments}

\subsection{Implementation Details}

\paragraph{Datasets} \mbox{}
For the training of our model, we utilized the publicly available MARIO dataset from \cite{TextDiffuser}, excluding MARIO-TMDB, and MARIO-OpenLibrary subsets as they are not publicly accessible. Upon removing any corrupted images, our final dataset comprised 7,249,449 image-caption pairs. In addition, to bolster our model's capability in generating a broader range of concepts (beyond text-centric images), we integrated 2,110,745 non-text images. These additional images, accompanied by text pairs, were selected based on a minimum predicted aesthetics score of 6.25, allowing for joint training to enhance overall performance.

\paragraph{Baselines} \mbox{}
We conducted quantitative comparisons of our SceneTextGen method against several leading approaches, including LDM\cite{stableDiffusion}, ControlNet\cite{controlNet}, TextDiffuser\cite{TextDiffuser}, GlyphControl\cite{yang2024glyphcontrol}, and DeepFloyd\cite{deepFloyd-if}, utilizing the publicly available code and pre-trained models for fairness. Notably, DeepFloyd is distinguished by its dual super-resolution modules, enabling it to produce high-resolution images at 1024x1024 pixels, in contrast to the 512x512 pixel images generated by the other models. For ControlNet comparisons, we employed Canny edge maps of printed text images created by the initial model stage of TextDiffuser as conditioning inputs. However, due to the unavailability of APIs, open-source code or checkpoints, we could not extend our comparative analysis to include Imagen\cite{Imagen}, eDiff-i\cite{ediffi}, or GlyphDraw\cite{GlyphDraw}.

\paragraph{Evaluation Criteria} \mbox{}
We assess text rendering quality using the MARIO-7M-Eval, MARIO-TMDB-Eval, and MARIO-OpenLibrary-Eval datasets. Our evaluation is twofold: firstly, through CLIPScore, which measures the cosine similarity between the image and text representations from CLIP; and secondly, via OCR Evaluation, which leverages existing OCR tools to detect and recognize text regions within the generated images. Metrics such as Average Precision, Average Recall, F1 Score, and Accuracy are employed to determine the presence of keywords in the generated images. 
During training, the input of the character encoder is the ground truth OCR texts. 
During the inference process, the character encoder receives as its input the text from captions provided by the user. These captions are enclosed in quotation marks as specified in the user's prompts, and the input of the CLIP text encoder is the full caption. For each generated and ground truth image pair, we utilize the easy-ocr library for OCR detection and recognition, followed by Hungarian Matching between the sets of texts, applying the Levenshtein distance (or edit distance) on the matched text pairs for the OCR evaluation.

\input{tables/OCR_scores}

\input{tables/CLIP_score}

\input{figures/qualitative}

\begin{algorithm}
\caption{T2I Visual Text Performance Evaluation}
\small
\begin{algorithmic}
\Function{Metrics}{$M$}
    \State $P \gets \Call{Precision}{M}$
    \State $R \gets \Call{Recall}{M}$
    \State $F \gets \Call{F1}{P, R}$
    \State $A \gets \Call{Accuracy}{M}$
    \State \Return $\{P, R, F, A\}$
\EndFunction
\end{algorithmic}
\small
\begin{algorithmic}[1]
\State $GT \gets \Call{OCR}{\text{Ground Truth Images}}$
\State $Gen \gets \Call{OCR}{\text{Generated Images}}$
\State $Scores \gets \text{an empty list}$
\For{each $pair$ in $\Call{Zip}{GT, Gen}$}
    \State $C \gets \Call{CostMat}{pair}$
    \State $M \gets \Call{Hungarian}{C}$
    \State $Scores \gets Scores \cup \{\Call{Metrics}{M}\}$
\EndFor
\State \Return $\Call{Average}{Scores}$
\end{algorithmic}
\end{algorithm}

\paragraph{Pseudo Code for OCR Performance Evaluation} \mbox{} 
The methodology described in Algorithm 1 demonstrates the steps taken to assess the performance of the text-to-image conversion. 
In the initial step, the algorithm applies OCR to both the ground truth and the generated images. This process results in two sets of text outputs, which are then paired for comparison. The Hungarian algorithm is employed here to find the optimal matching between elements (words) of these two sets, minimizing the overall difference between the matched pairs. This is crucial for an objective and accurate comparison. For each matched pair, we calculate a cost matrix, which serves as the input for the Hungarian algorithm. The output of this step is a matching matrix \(M\) which represents the best possible alignment between the text elements in the ground truth and the generated images. Subsequently, the algorithm computes key performance metrics for each pair: Precision, Recall, F1 Score, and Accuracy.  Precision focuses on the accuracy of the replicated text, recall measures the completeness, F1 score provides a balance between precision and recall, and accuracy gives an overall effectiveness of the text replication. Finally, the algorithm averages these scores across all image pairs to provide an overall performance evaluation of the text-to-image conversion process.

\subsection{Quantitative Results}
Our experimental analysis in Tab. \ref{table:ocr} provides a direct comparison of the OCR based recognition scores among various models as measured in terms of Precision, Recall, F1 scores, and Accuracy. Our results indicate that SceneTextGen consistently outperforms competing models in most metrics. Latent Diffusion Model, lacking a sophisticated mechanism for text comprehension, typically under-performs, leading to lower OCR scores. In contrast, DeepFloyd\cite{deepFloyd-if} incorporates a T5 encoder which aids in textual understanding, thereby enhancing the quality of the generated text. However, its performance is still limited due to an insufficient character-level understanding.

ControlNet\cite{controlNet}, TextDiffuser\cite{TextDiffuser}, and GlyphControl\cite{yang2024glyphcontrol}, which utilize predefined text layouts or spatial information, show mixed results. While the explicit introduction of text information allows ControlNet to achieve high OCR scores, the resulting text often appears artificial and lacks seamless integration within the images (see Fig. \ref{fig:comparison}).

\paragraph{Cross-dataset Generalization Ability} \mbox{}
As demonstrated in Tab. \ref{table:ocr}, SceneTextGen-7M, despite being trained solely on the MARIO-7M dataset, exhibits strong generalization capabilities. It maintains robust OCR scores across evaluation sets from both the TMDB and OpenLibrary datasets, underscoring its adaptability and the efficacy of its training methodology.

\subsection{Measuring Font Style Diversity}
To quantitatively assess the diversity of font styles generated by SceneTextGen, we utilized a pretrained VGG-based font recognition model trained on synthesized text images. This approach involved first extracting text image patches using a pretrained Optical Character Recognition (OCR) model. These patches were then processed through the font recognition model to retrieve features from the penultimate layer. By applying t-SNE for dimensionality reduction, we visualized the feature space to examine the proximity and diversity of text generated by different methods.

\begin{figure}[!htbp]
\centering
    \includegraphics[width=0.8\linewidth]{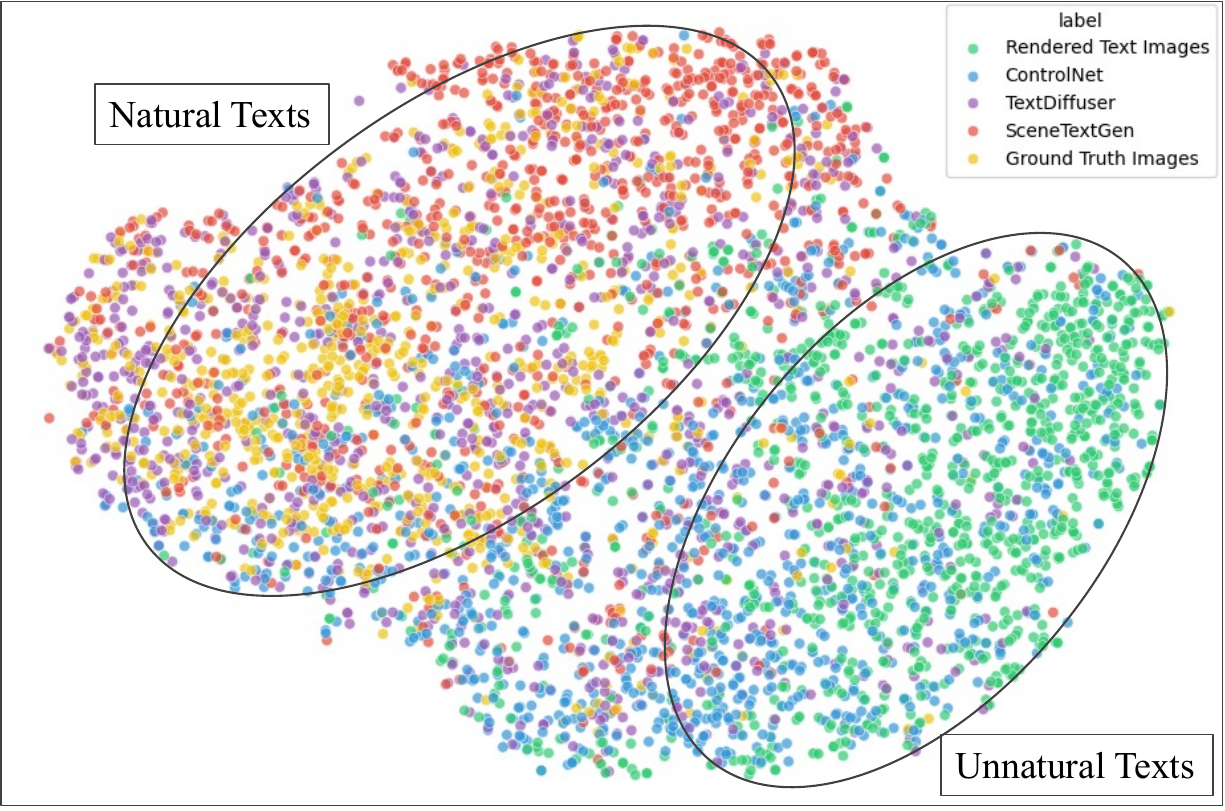}
    \caption{t-SNE representation of text region embeddings derived from the penultimate layer features of a font recognition model. }
    \label{fig:tSNE}
\end{figure}

\begin{figure}[!htbp]
\centering
    \includegraphics[width=1\linewidth]{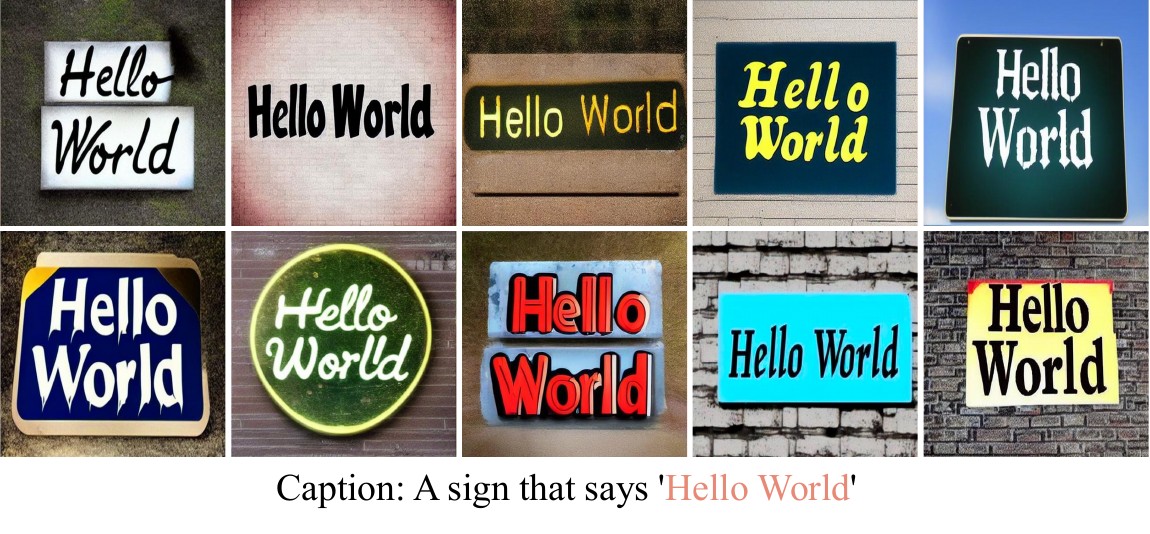}
\caption{Diversity of font styles. SceneTextGen is able to generate visually appealing and diverse font styles and layouts for text, without any style or layout prompts.}
\label{fig:diversity}
\end{figure}

\begin{figure}[!htbp]
\centering
    \includegraphics[width=1.0\linewidth]{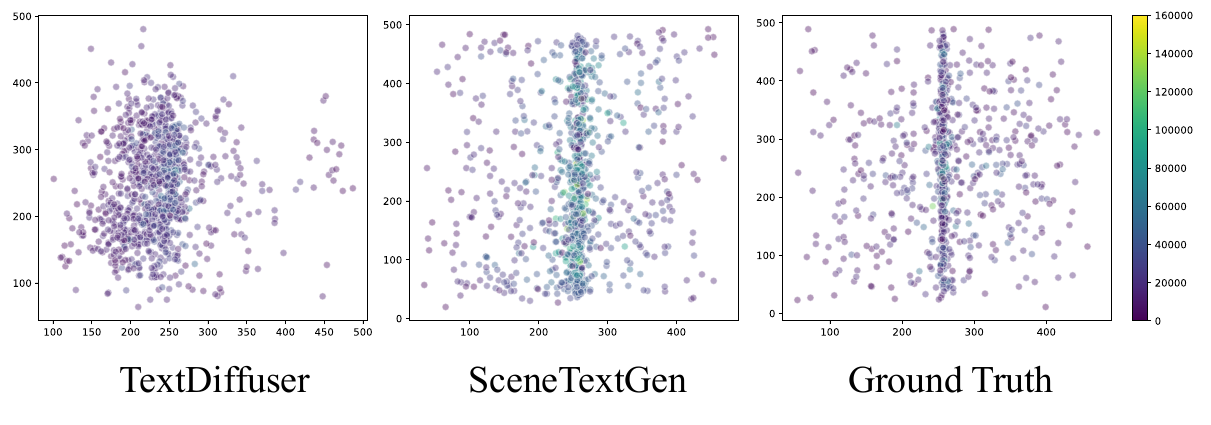}
\caption{Distribution of visual text coordinates in the generated image w.r.t. the area of each box. As expected from real-life observations, the visual text in the ground truth images tends to be located at the center of each image.}
\label{fig:distribution}
\end{figure}

As illustrated in Fig. \ref{fig:tSNE}, the rendered text images — which serve as a baseline — were created by printing text onto a white canvas using the layout generator from \cite{TextDiffuser} with the default 'Arial' font. This process establishes a reference for the feature distribution of 'Artificial Texts.' ControlNet \cite{controlNet}, constrained by predefined text canny edges, exhibits a feature distribution closely aligned with the rendered text images. This proximity suggests its limitations in integrating text seamlessly within images, a finding corroborated by the qualitative results shown in Fig. \ref{fig:comparison}. TextDiffuser \cite{TextDiffuser}, although performing well in most instances, still shares some feature space with ControlNet, indicating occasional production of artificial or unnatural texts. In contrast, SceneTextGen — operating independently of any predefined canny edges or layouts — demonstrates a distinct distribution with minimal overlap with the 'rendered text images,' signifying its robustness in generating naturalistic text within the context of scene images. Example visualizations are in Fig. \ref{fig:diversity}. In addition, to better understand the text layout given by each model, we also show in Fig. \ref{fig:distribution} the overall distribution of visual text in the generated images and ground truth images.

\subsection{Weight of Each Loss}

In this section, we present two ablation studies to evaluate the impact of different loss function configurations on OCR performance. Table \ref{tab:weight_ablation_study} explores the effects of using combinations of word-level and character-level losses. Table \ref{tab:lambda_ablation_study} examines the impact of varying the weights of these loss functions. In both studies, we assess the Average Precision (AP) and Accuracy (AC) scores to understand how these configurations influence the model's ability to accurately recognize and generate text in images.

\input{tables/ablation}

\subsection{Limitations}

\begin{figure}[!htbp]
    \centering
    \includegraphics[width=1\linewidth]{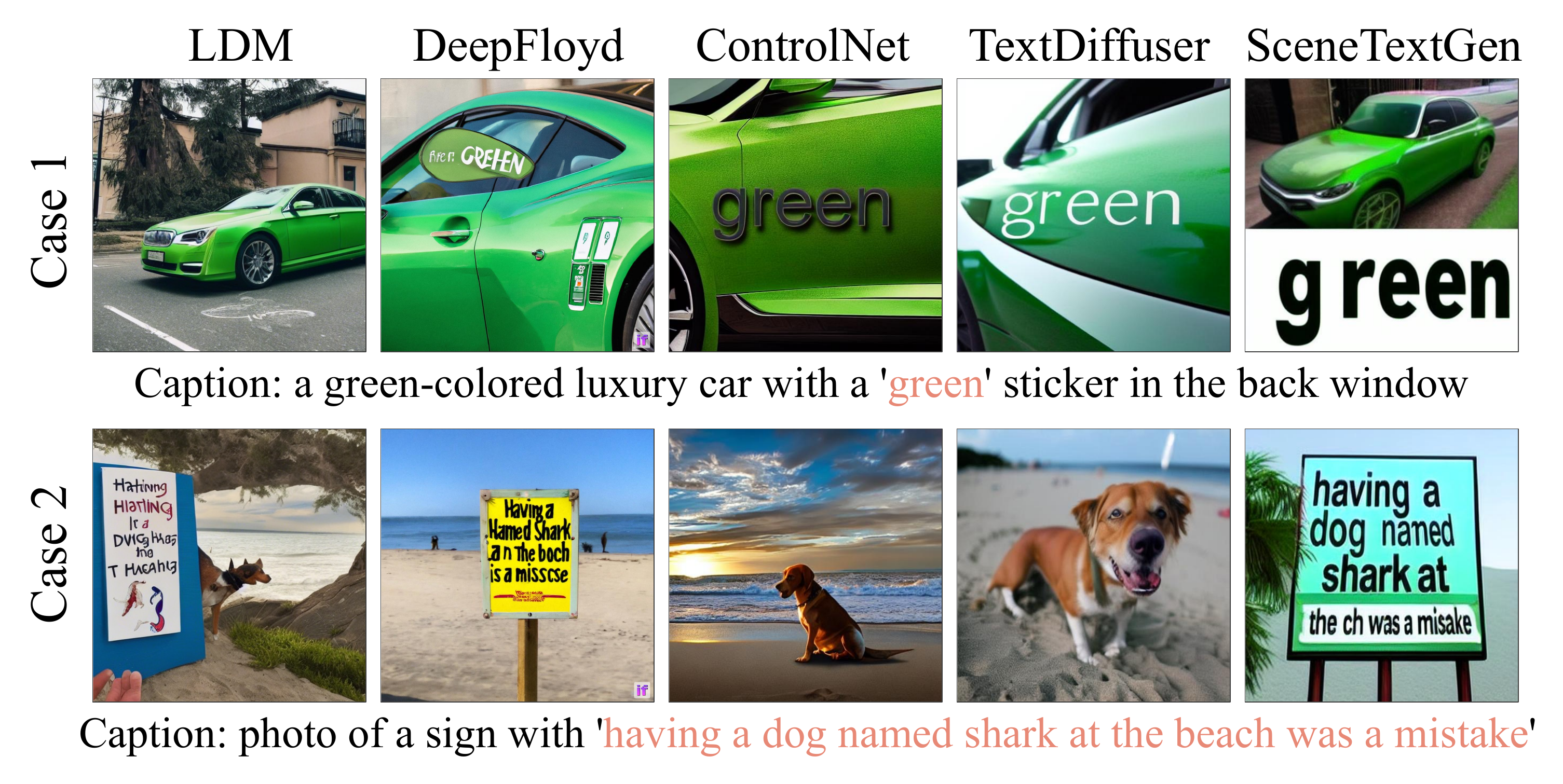}
    \vspace*{-2em}
    \caption{Failure cases in complex scene interpretation and processing of lengthy prompts for text.}
    \label{fig:failure_general_length}
\end{figure}

While SceneTextGen demonstrates superior performance in scene text image generation, challenges remain. Limited scene complexity: Models struggle to accurately generate complex elements in conjunction with text, such as a car with a 'green' sticker specifically in the back window.(Fig.~\ref{fig:failure_general_length}, Case 1). Long text handling: Text accuracy and coherence decrease with increasing length (Fig.~\ref{fig:failure_general_length}, Case 2). These areas require further development.

%% file: tables/OCR_scores.tex
\setlength{\tabcolsep}{4pt}
\begin{table*}[!htbp]
\centering
{\footnotesize 
\begin{tabular}{l c cccc cccc cccc}
\toprule
\multirow{2}{*}{} & \multicolumn{1}{c}{\multirow{2}{*}{\shortstack[c]{Pre-defined \\Text Layout}}}  & \multicolumn{4}{c}{MARIO-7M} & \multicolumn{4}{c}{TMDB} & \multicolumn{4}{c}{OpenLibrary}  \\
\cmidrule(lr){3-6} \cmidrule(lr){7-10} \cmidrule(lr){11-14}
  & & AP($\uparrow$) & AR($\uparrow$) & F1($\uparrow$) & AC($\uparrow$) & AP($\uparrow$) & AR($\uparrow$) & F1($\uparrow$) & AC($\uparrow$) & AP($\uparrow$) & AR($\uparrow$) & F1($\uparrow$) & AC($\uparrow$)  \\

\toprule
  TextDiffuser-10M \cite{TextDiffuser} & \checkmark &  0.6135  & 0.4289  & 0.4683  & 0.3425 &  0.4401 & 0.4243  & 0.3994  & 0.2889   & 0.5617 &  0.3816  & 0.4242   & 0.2916  \\
  GlyphControl-10M* \cite{yang2024glyphcontrol} & \checkmark &  0.5075  & 0.4118  & 0.4140  & 0.2883 &  0.3635 & 0.4082  & 0.3457  & 0.2362  & 0.4734  &  0.3762  & 0.3836  & 0.2534  \\
\toprule
  Latent Diffusion Model  &            &  0.1482  & 0.1690  & 0.1296  & 0.0753 & 0.1717  & 0.2579  & 0.1703  & 0.0974  & 0.1911  &  0.2873  & 0.1923  & 0.1106  \\
  DeepFloyd \cite{deepFloyd-if}       &            &  0.2467  & 0.2788  & 0.2206  & 0.1366 &  0.2284 & 0.3738  & 0.2449  & 0.1500  & 0.2555  &  0.3910  & 0.2635  & 0.1610  \\
  ControlNet \cite{controlNet}     & \checkmark &  0.5102  & \textbf{0.4444}  & 0.4238  & 0.2981 &  0.3075 & 0.4667  & 0.3284  & 0.2194  & 0.4050  &  0.4273  & 0.3690  & 0.2415  \\
  TextDiffuser-7M \cite{TextDiffuser} & \checkmark &  0.4778  & 0.3447  & 0.3682  & 0.2512 &  0.3198 & 0.3362  & 0.2961  & 0.1930  & \textbf{0.4257}  &  0.3146  & 0.3318  & 0.2108  \\
  SceneTextGen-7M(Ours) &       &  \textbf{0.5274}  & 0.4420  & \textbf{0.4424}  & \textbf{0.3088} &  \textbf{0.3813} & \textbf{0.4716}  & \textbf{0.3790}  & \textbf{0.2602}  & 0.4136  &  \textbf{0.4519}  & \textbf{0.3945}  & \textbf{0.2571}  \\
\bottomrule
\end{tabular}
}
\caption{Comparative analysis of OCR based text recognition scores across different models. Note, 7M denotes models that were trained on the MARIO-7M dataset (7 million images with texts). In contrast, the TextDiffuser-10M category includes models trained on an expanded dataset collection that encompasses MARIO-7M, MARIO-TMDB, and MARIO-OpenLibrary. * denotes LAION-Glyph dataset.}
\label{table:ocr}
\end{table*}
\setlength{\tabcolsep}{1.4pt}

%% file: tables/CLIP_score.tex
\setlength{\tabcolsep}{4pt}
\begin{table}[!htbp]
\centering
{\footnotesize 
\begin{tabular}{l c c c c}
\toprule
  & \shortstack[c]{Pre-defined \\Text Layout} & CLIP Score($\uparrow$) \\
\toprule
  TextDiffuser-10M \cite{TextDiffuser} &  \checkmark  &  0.3436   \\
  GlyphControl-10M* \cite{yang2024glyphcontrol}  &  \checkmark  &  0.3450    \\
\toprule
  Latent Diffusion Model &              &  0.3015   \\
  DeepFloyd \cite{deepFloyd-if}        &              &  0.3267   \\
  ControlNet \cite{controlNet}       &  \checkmark  &  0.3424   \\
  TextDiffuser-7M \cite{TextDiffuser}  &  \checkmark  &  0.3385    \\
  SceneTextGen-7M(Ours)  &         &  \textbf{0.3455}   \\
\bottomrule
\end{tabular}
}
\caption{Comparison of CLIP scores reflecting the overall image quality generated by various models. SceneTextGen-7M achieves the highest CLIP score, indicating superior text-image alignment without relying on text layouts. * denotes LAION-Glyph dataset.}
\label{table:clip_score}
\end{table}
\setlength{\tabcolsep}{1.4pt}

%% file: figures/qualitative.tex
\begin{figure*}[!htbp]\centering
    \includegraphics[width=.91\linewidth]{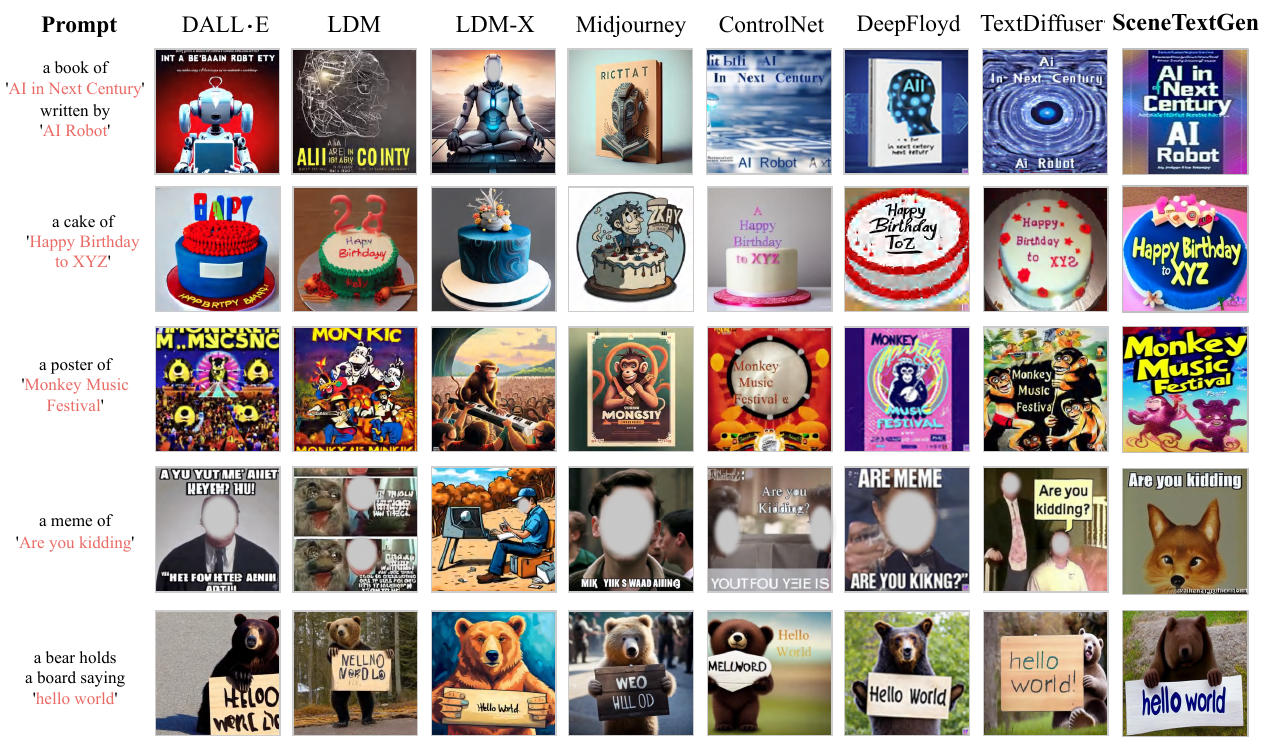}
    \caption{
Comparative visualization of generated images. We present a side-by-side comparison of images generated from the same text prompt across different existing methods (with results generated by ~\cite{TextDiffuser} as a proxy. Human faces are blurred for ethical considerations.). Each row corresponds to a unique prompt, showcasing the visual quality, text clarity, and contextual coherence achieved by each method.}
    \label{fig:comparison}
\end{figure*}

%% file: tables/ablation.tex
\begin{table}[!htbp]
\centering
\begin{minipage}{.46\linewidth}
\centering\footnotesize
\begin{tabular}{llll}
\hline
L$_{\textrm{word}}$ & L$_{\textrm{char}}$ & \textbf{AP} & \textbf{AC} \\ \hline
\checkmark          &                                  & 0.5082      & \textbf{0.3109}      \\
               & \checkmark      & 0.5083      & 0.3035      \\
\checkmark          & \checkmark                        & \textbf{0.5274}      & 0.3088      \\ \hline
\end{tabular}
\caption{Effects of loss function combinations on the MARIO-7M-Eval dataset}
\label{tab:weight_ablation_study}
\end{minipage}\hfill
\begin{minipage}{.46\linewidth}
\centering\footnotesize
\begin{tabular}{llll}
\hline
W$_{\textrm{lam}}$ & C$_{\textrm{lam}}$ & \textbf{AP} & \textbf{AC} \\ \hline
1    & 2    & 0.4107  & 0.2782 \\
0.1  & 0.2  & 0.4587  & 0.2895 \\
0.01 & 0.02 & \textbf{0.5274}  & \textbf{0.3088} \\
\hline
\end{tabular}
\caption{Impact of varying loss weights on the MARIO-7M-eval dataset}
\label{tab:lambda_ablation_study}
\end{minipage}
\end{table}

%% file: sections/5_discussion_and_conclusion.tex
\section{Conclusion}

SceneTextGen, incorporating a character-level encoder and hierarchical text integration, offers advancements in scene text image generation. Despite improved text rendering, limitations arise in generating complex visuals and handling lengthy text. These challenges highlight the ongoing difficulty in reconciling textual accuracy with broader image synthesis. This work furthers our understanding of text-image generation, paving the way for future exploration.

\paragraph{Acknowledgements} This research project has been partially funded by research grants  to Dimitris N. Metaxas through NSF: 2310966, 2235405, 2212301, 2003874, and FA9550-23-1-0417.

\newpage